\documentclass[conference]{IEEEtran}
\IEEEoverridecommandlockouts
\usepackage[caption=false,font=footnotesize]{subfig}
\usepackage[rightcaption]{sidecap}
\usepackage{amsmath,amssymb,amsfonts,bm}
\usepackage{algorithm,algpseudocode}
\usepackage{graphicx}
\usepackage[english]{babel}
\usepackage{url}

\usepackage[square,numbers]{natbib}
\usepackage{xcolor}
\def\BibTeX{{\rm B\kern-.05em{\sc i\kern-.025em b}\kern-.08em
    T\kern-.1667em\lower.7ex\hbox{E}\kern-.125emX}}
\begin{document}

\title{Detecting inter-sectional accuracy differences in driver drowsiness detection algorithms.\\
}

\author{\IEEEauthorblockN{ Mkhuseli Ngxande}
\IEEEauthorblockA{\textit{CSIR Defence, Peace, Safety and } \\
\textit{Security} \\
\textit{Optronic Sensor Systems, }\\
Pretoria, South Africa \\
Email: mngxande@csir.co.za}
\and
\IEEEauthorblockN{Jules-Raymond Tapamo }
\IEEEauthorblockA{\textit{School of Electrical, Electronic and } \\
\textit{Computer Engineering}\\
\textit{University of Kwa-Zulu Natal,}\\
Durban, South Africa  \\
Email: tapamoj@ukzn.ac.za}
\and
\IEEEauthorblockN{ Michael Burke}
\IEEEauthorblockA{\textit{Mobile Intelligent Autonomous Systems} \\
\textit{Modelling and Digital Sciences}\\
\textit{Council for Scientific and Industrial Research}\\
Pretoria, South Africa \\
Email: michaelburke@ieee.org}

}
\maketitle

\begin{abstract}
Convolutional Neural Networks (CNNs) have been used successfully across a broad range of areas including data mining, object detection, and in business. The dominance of CNNs follows a breakthrough by Alex Krizhevsky which showed improvements by dramatically reducing the error rate obtained in a general image classification task from 26.2\% to 15.4\%. In road safety, CNNs have been applied widely to the detection of traffic signs, obstacle detection, and lane departure checking. In addition, CNNs have been used in data mining systems that monitor driving patterns and recommend rest breaks when appropriate. This paper presents a driver drowsiness detection system and shows that there are potential social challenges regarding the application of these techniques, by highlighting problems in detecting dark-skinned driver’s faces. This is a particularly important challenge in African contexts, where there are more dark-skinned drivers. Unfortunately, publicly available datasets are often captured in different cultural contexts, and therefore do not cover all ethnicities, which can lead to false detections or racially biased models. This work evaluates the performance obtained when training convolutional neural network models on commonly used driver drowsiness detection datasets and testing on datasets specifically chosen for broader representation. Results show that models trained using publicly available datasets suffer extensively from over-fitting, and can exhibit racial bias, as shown by testing on a more representative dataset. We propose a novel visualisation technique that can assist in identifying groups of people where there might be the potential of discrimination, using Principal Component Analysis (PCA) to produce a grid of faces sorted by similarity, and combining these with a model accuracy overlay.  
\end{abstract}


\section{Introduction}
The convolutional neural network (CNN) has rapidly gained popularity in many social aspects and has been applied across a range of areas, including self-driving cars, collision detection, identification of criminal activities, and to aid the granting of bank loans. Historically, these tasks were generally performed by humans, but the advancement of machine learning is leading to the automation of these processes \cite{o2016weapons}. CNNs are a multistage mechanism that learns data representations in order to fulfil a specific goal. However, these can suffer from challenges with regard to generalisation. For example, a system that is trained to detect road lanes in an urban environment and then deployed in rural areas could lead to false detections. Furthermore, driver drowsiness detection systems predominately trained on a certain race or ethnicity may not perform well when tested across multiple races. This can potentially result in algorithmic discrimination if the trained model is unable to handle differing skin complexions and facial features \cite{bolukbasi2016man}. This raises concerns in African contexts, where many cars with driver drowsiness detection systems are imported \cite{NationalAssoc}.

For example, the majority (80.8\%) of South African citizens identify as black nationals \cite{StatisticsSouthAfrica2017}. Deploying a system that is trained in different contexts, for example, using a dataset captured in Asia, could result in failure if trained models learn to use skin complexion for decision-making. The taxi industry dominates public transportation used in South Africa. Statistics South Africa \cite{Africa2016} report that about 76\% of citizens in the country use public transportation to get to their destinations, with private minibus taxis a primary mode of transport (51.0\%), followed by busses at 18.1 \% and trains at 7.6\%.

The alarming statistics of road accidents in South Africa has led to the investigation of technologies to reduce these high numbers of accidents. A Statistics South Africa report showed that in 2015, there was a 2\% increase in mortality over the 2014 financial year, with about 12944 deaths caused by accidents \cite{RTMC2017}. Furthermore, in 2016 there were about 14 071 deaths, which was a 9\% increase over 2015 \cite{RTMC2017}.\\

Drowsiness detection systems that are currently implemented are typically available only in high-end vehicles, which disadvantages citizens using public transport. As a result, a number of researchers have aimed to develop similar systems on mobile phones, which are more easily accessible \cite{J2013}. In addition to easily accessible systems, researchers also focused on augmenting vehicle control units with machine learning techniques to help reduce road accidents \cite{8500333,8506402}.  However, if a large benchmark dataset that is representative of all race ethnicities is not used for training, systems like these can easily fail. Sikander and Anwar conducted a review of existing technologies for detection of fatigue in drivers, where various techniques and features were examined \cite{8482470}. Of the 23 fatigue detection systems reviewed here, 12 relied on machine learning techniques.  Moreover, in \cite{ngxande2017driver} it is shown that CNNs tended to outperform other technologies for driver drowsiness detection. However, the authors note that there is no large benchmarking dataset covering a wide range of ethnicities with which to conclusively test the efficiency of CNNs against other technologies.

This paper aims to highlight the challenges of using unrepresentative images to train vision-based driver drowsiness detection systems. In this article, we use a range of pre-trained convolutional neural networks, modifying the last layers by retraining these on a number of popular drowsiness detection datasets including the ULg Multimodality Drowsiness Dataset (DROZY), the National Tsinghua University Drowsy Driver Detection database (NTHU-DDD), and the Closed Eyes in the Wild dataset (CEW). We test models trained using these datasets on a test set more suited to South African contexts. Results show that the three evaluated datasets produce high drowsiness detection accuracies when tested on held out portions of the original datasets, but that the accuracies obtained decreased substantially when these models were evaluated using the more representative South African test set. This decrease in performance is due to models overfitting. Overfitting models can be a particular challenge, as it can be difficult to establish where models are failing to generalise. This work introduces a new visualisation technique to identify potential population groups for whom additional training data may be required, so as to rectify the problem of unrepresentative models in driver drowsiness detection systems.\\

This paper is structured as follows. Section II provides an overview of related work, which is followed by a discussion on algorithmic bias and convolutional neural networks. This is followed by an introduction of the proposed visualisation technique in Section III, and a description of the experimental methods, including the architectures and datasets investigated in Section IV. Finally, results and the conclusions are provided in Section V and VI respectively.

\section{BACKGROUND AND RELATED WORK}

This section briefly summarises previous approaches to driver drowsiness detection and existing benchmarking datasets used for testing these algorithms. Related work and advances in convolutional neural network architectures are also discussed.

\subsection{DRIVER DROWSINESS DETECTION SYSTEMS}

A number of drowsiness detection systems rely on convolutional neural networks. Sanghyuk et al. \cite{park2016driver} proposed a deep architecture called deep drowsiness detections (DDD). This architecture consists of three deep convolutional neural networks including AlexNet \cite{krizhevsky2012imagenet}, VGGNet-FaceNet \cite{parkhi2015deep}, and FlowImageNet \cite{donahue2015long}. The output of these networks is concatenated and fed into a softmax classification layer for drowsiness detection. The DDD system was tested on the NTHU-drowsy driver detection dataset, but the authors noted that the NTHU-drowsy lacked reliable ground truth labeling, which led them to use a substitute evaluation dataset for testing. The authors also noted that there was a lack of previously benchmarked datasets to compare with the publicly available NTHU-drowsy dataset.

Reddy et al. \cite{reddy2017real} proposed a compressed deep neural network model that can be deployed on an embedded board. The authors note that for their focus, the NTHU-drowsy dataset had an unsuitable capture angle and inappropriate class labels. In addition, the authors noted that the DROZY dataset was also unsuitable because the images contain sensor patches attached to a subject’s face, which could interfere with the results obtained. Their solution was to use a custom dataset and compare the efficacy of their approach to a number of convolutional neural network architectures, including faster RCNN, VGG-16, and AlexNet.

 Lyu et al. \cite{lyu2018long} proposed a sequential multi-granularity deep framework for detection of driver drowsiness. This framework consists of two components, a multi-granularity CNN and a deep long-short-term memory network (deep LSTM). A contribution of this work was to utilise a group of parallel CNN extractors. The deep LSTM was applied on facial representations to identify long-term features of drowsiness over a sequence of frames. The model was evaluated on the NTHU-drowsy dataset in addition to a new dataset named Forward Instant Driver Drowsiness Detection (FI-DDD). The FI-DDD is a re-labeled NTHU-drowsy dataset, as the authors note that it is difficult to locate drowsy states temporally with high precision using the NTHU-drowsy labels. 

Following a different approach, Dwivedi et al. \cite{dwivedi2014drowsy} introduced a more diverse dataset that includes persons with different skin tones, eye shapes and eye sizes. This dataset was used to test a CNN with a final softmax classification layer, but unfortunately, the dataset is not publicly available for comparison. \\

A recent study by Kim et al. proposed a deep CNN based on the classification of opened and closed eyes using a visible light camera sensor \cite{kim2017study}. They used the ZJU eye blink dataset in addition to their own dataset collected for performance analysis. Here, the ResNet-50 \cite{He2017} architecture was adopted, with a modified fully connected layer. The system outperformed AlexNet \cite{krizhevsky2012imagenet}, GoogleNet \cite{Szegedy_2015_CVPR}, VGGFace fine-turning  \cite{parkhi2015deep}, and HOG-SVM \cite{pauly2015detection}.

\subsection{ALGORITHMIC BIAS}

It is clear that a number of modern drowsiness detection systems rely on convolutional neural networks, and many of these models are trained and tested on only a few datasets. These datasets do not always cover a wide range of different races and ethnicities with varying facial features. As a result, these systems are vulnerable to problems regarding algorithmic bias. This paper evaluates the efficacy of the NTHU-drowsy, DROZY, and CEW datasets in a South African context, where racial bias is likely to have a significant impact, using a more representative dataset.

Unfortunately, the road safety community is not the only field that is affected by algorithmic bias caused by using unrepresentative datasets for training. Buolamwini \cite{buolamwini2017gender} have documented extensive algorithmic bias in face detection systems, which fail on faces with darker skin-tones, while Renda et. al have highlighted bias in predictive policing \cite{renda2018ethics}, by showing that a system called PredoPol used to send police to crime hotspots tends to send police to areas where there are large numbers of dark-skinned people or Muslims. Wen et al. \cite{wen2015face} analysed a face spoof detection algorithm, designed to recognise fake faces using image distortion analysis and reported that most current systems mis-classify individuals with dark-skinned faces as spoof attacks.\\

Furthermore, Brauneis and Goodman added to the discussion of how to deploy AI-based systems, evaluating a number of scenarios where dark-skinned people could be mis-classified \cite{Brauneis2018}. Zou and Schiebinger \cite{Zou2018} shows that there is often bias in machine learning algorithms, which can be caused by a variety of factors including imbalanced training datasets, representation of the datasets and also algorithms themselves. They suggest that datasets should include information on how they were collected and the demographics of participants therein using meta-data. However, this process is also problematic, as it requires the classification of people into different ethnic groups or categories, which is itself a subjective and questionable task. In an attempt to address this, we propose a visualisation technique that can identify groups or individuals on whom algorithms fail, without the need for pre-classification or meta-data.  

\subsection{CONVOLUTIONAL NEURAL NETWORKS}

CNNs have dominated many computer vision tasks since the breakthrough shown by Alex Krizhevsky in the ImageNet Large-Scale Visual Recognition Challenge (ILSVRC) in 2012 \cite{krizhevsky2012imagenet}. Convolutional neural networks are a collection of sequentially stacked layers, which typically consist of convolution, pooling, and fully connected layers, with model parameters trained using gradient descent. The convolutional layer takes in a three-dimensional image tensor, comprising $d$ channels of feature maps, sized $h \times w$ pixels. Here, $h$ denotes the height and $w$ the width of the input image tensor. Convolutional layers extract low level features from an input tensor $I$ by means of a convolution with a two-dimensional kernel $K$,
\begin{IEEEeqnarray}{lCl}
c(i,j) &=& (K \ast I)(i,j)\nonumber\\
&=&\sum_{m}\sum_{n}I(i-m,i-n)K(m,n)+ b_{m,n} \label{eq1}
\end{IEEEeqnarray}
where $b_{m,n}$ is a bias parameter, and $i,j$ denote the coordinates of a feature map pixel. After convolution, an activation function is applied to introduce non-linearity and produce an output feature map $\mathbf{a}$, comprising elements, $a(i,j) = f(c(i,j))$. A number of activation functions can be used, but the ReLU function, 
\begin{equation}
f(c)= \begin{cases} \label{eq2}
    c,&\text{if } c > 0 ,\\    
    0,&\text{if } c \leq 0,
\end{cases}
\end{equation}
is typically used for convolutional neural networks in order to avoid vanishing gradients in deeper networks. 

Pooling layers are often applied after activations. Here, downsampling is applied to reduce the dimensionality of the image by a grouping operation over activations in small spatial regions of the input image. For example, max pooling returns the maximum value of the input region. Max pooling is also used to control over-fitting. Downsampling can also be achieved by using a convolutional layer with larger filtering strides. 


A fully connected output layer is typically the final layer in a convolutional neural network, producing a network output: 
\begin{equation}
\mathbf{Z} = \mathbf{W}\mathbf{a} + \mathbf{b}\label{eq3}
\end{equation}
where $\mathbf{W}$ is a weight matrix and $\mathbf{a}$ the input from the previous layer. This is typically followed by a final activation layer. The activation function that is used for this paper is the sigmoid function, which is commonly used for binary classification tasks.

Convolutional neural networks are trained using gradient descent to find model parameters that minimise some loss. For binary classification tasks like drowsiness detection, the binary cross entropy loss given by 
\begin{equation}
J = - \sum ^{r}_{i=1}y_{i}lo\textit{g}(o_{i})\label{eq4},
\end{equation}
is typically applied, where $y$ is a vector of one-hot encoded labels, and $o$ is the output probability produced by the final sigmoid layer. This loss function is typically minimised using stochastic gradient descent schemes to adjust model weight and bias parameters, with the Adam optimiser \cite{Kingma2017} often favoured.

A number of convolutional neural network architectures have been developed. The following section briefly highlights some of the improvements made thus far, by highlighting results in the ImageNet Large-Scale Visual Recognition Challenge (ILSVRC):

\begin{LaTeXdescription}
\item[ZFNet (2013)]In 2013 the winners of the ILSVRC were Matthew Zeiler and Rob Fergus from New York University \cite{zeiler2014visualizing}. Their model achieved an 11.2\% error rate, improving upon 2012 error rate of 15.4\% obtained by AlexNet. Although ZFNet is similar to AlexNet, ZFNet was modified by introducing a deconvNet and decreasing filter sizes from 11 x 11 pixels to 7 x 7 pixels.

\item[GoogleNet (2014)]GoogleNet is a 22 layer convolutional neural network that won the 2014 ILSVRC challenge. The novelty of this work was the introduction of the Inception module \cite{Szegedy_2015_CVPR}. This module aims to reduce computational costs while increasing the width and depth of the network. The inception module has been extended several times with recent iterations including Inception-V3 \cite{szegedy2016rethinking} and Inception-V4 \cite{szegedy2017inception} models.

\item[VGGNet (2014)]The VGGNet developed by Karen Simonyan and Andrew Zisserman consists of two versions (VGG16 and VGG19 models), and took second place in the ILSVRC 2014 challenge \cite{simonyan2014very}. The two network architectures have depths of 16 and 19 layers respectively. VGGNet decreased the filter sizes of ZFNet to 3x3 with the motivation that these smaller filter sizes are capable of gathering more information from input images.

\item[ResNet (2015)] This network, developed by Microsoft Research Asia, won the 2015 ILSVRC with an error rate of 3.6\% \cite{He2017}.  This model uses a residual learning framework that aims to simplify the training of deeper networks and yield higher accuracy. This network consists of 152 layers and was extended to 1001 layers on CIFAR-10, achieving an error rate of 4.62\% \cite{he2016identity}.
\end{LaTeXdescription}

It is clear that there is a trend of increasing the depth of the network, producing increasing performance, while reducing computational costs. However, these trends can make convolutional neural networks more vulnerable to over-fitting. Strategies for avoiding over-fitting include Batch Normalization \cite{ioffe2015batch} and Dropout \cite{srivastava2014dropout}.

\section{VISUALISING CONVOLUTIONAL NEURAL NETWORKS}

Visualisation is a commonly used technique to interpret trained convolutional neural networks, so as to improve the architecture or identify model failures. For example, saliency  visualisation helps to identify which image areas contribute strongly to the network output. This technique was introduced by Simonyan et al. who presented two approaches to visualise what a neural network learns \cite{simonyan2013deep}. Here, the gradient of the class score with respect to image pixels is computed to determine the contribution of each pixel to the final output prediction \cite{selvaraju2017grad}. Many previous approaches \cite{zeiler2014visualizing,springenberg2014striving,fong2017interpretable,kindermans2017reliability,kahng2018cti} tried to come up with visualisation solutions to better understand what each layer learns, but none of these directly address the challenge of potential racial bias in machine learning.

In this work, we introduce a visualisation technique building on Principal Component Analysis (PCA) to help identify population groups where models fail to perform well. Here, we use PCA to project images onto a 2-dimensional grid such that images are located near other images of similar appearance. PCA is a dimension reduction technique that can be used to compress a large number of features to a smaller number while retaining dominant information \cite{Smith2002}. This is done by transforming data into an orthogonal subspace where axes (Principal components) align with the directions of maximum variance in the data. In this work we use singular value decomposition (SVD) to perform PCA.  Let \textbf{X} be the matrix of images, formed by reshaping images $\textbf{x}_{i}$ into row vectors (where $ \textit{i} = 1 ... \textit{N}$, and \textit{N} is the number of images in the dataset) and stacking these vertically to form an \textit{N} x \textit{P} matrix. Here, \textit{P} denotes the number of pixels in each image. PCA starts by mean centering the matrix of images, which is accomplished by subtracting the mean image
$ \mu_{i} =\dfrac{1}{N} \sum ^{N}_{i}\textit{\textbf{X}}_{i} $ from each 1 x \textit{P}  dimensional row vector, $\textbf{X}_{i}$ in the matrix of images,\begin{center}
$ \mathbf{\hat{X}}_{i} = \textbf{X}_{i}-\mu_{i}  $
\end{center}
The mean centred matrix  \textit{N} x \textit{P}  dimensional matrix of images $\hat{\textbf{X}} $ is then decomposed using singular value decomposition (SVD),\begin{center}
$ \mathbf{\hat{X}} = \mathbf{U}\mathbf{\Sigma}\mathbf{V}^\text{T}$
\end{center}
Here, $\mathbf{U}$ is a $ P \times N$ unitary matrix, $\mathbf{V}$ is a $P \times P$ unitary matrix  and $\mathbf{\Sigma}$ is a diagonal matrix comprising the singular values of $\mathbf{\hat{X}}$ in decreasing order \cite{Stewart}. A reduced dimensional representation of $\mathbf{\hat{X}}$ can be obtained by discarding columns of $\mathbf{U}$ and $\mathbf{V}$,\begin{center}
$ \mathbf{\hat{X}} \approx \mathbf{U}_{0:j}\mathbf{\Sigma}_{0:j,0:j}\mathbf{V}_{0:j}^\text{T}$
\end{center}
Here, $j$ denotes the number of columns retained. As shown above, PCA can project data into a low dimensional coordinate system, with axes provided by the columns of $\mathbf{U}_{0:j}$, and data coordinates given by $\mathbf{V}_{0:j}$. 

\begin{figure}
\centering
\includegraphics [width=0.49\textwidth,height=6cm]{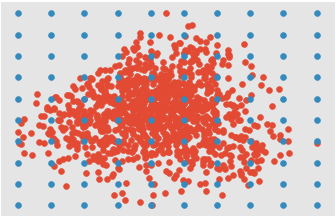}
\caption{The figure shows 2-dimensional grid projection coordinates obtained after applying the linear PCA transformation into 2-dimensional subspace. A uniformly spaced grid is placed over the projected image coordinates, and images are assigned a grid position by finding the closest image coordinate to each grid position, ensuring each image can only be used once. The blue dots represent grid position and the red dots represent PCA projections. \label{grid}}
\end{figure}

\begin{figure}
\centering
\includegraphics[width=0.49\textwidth,height=9cm]{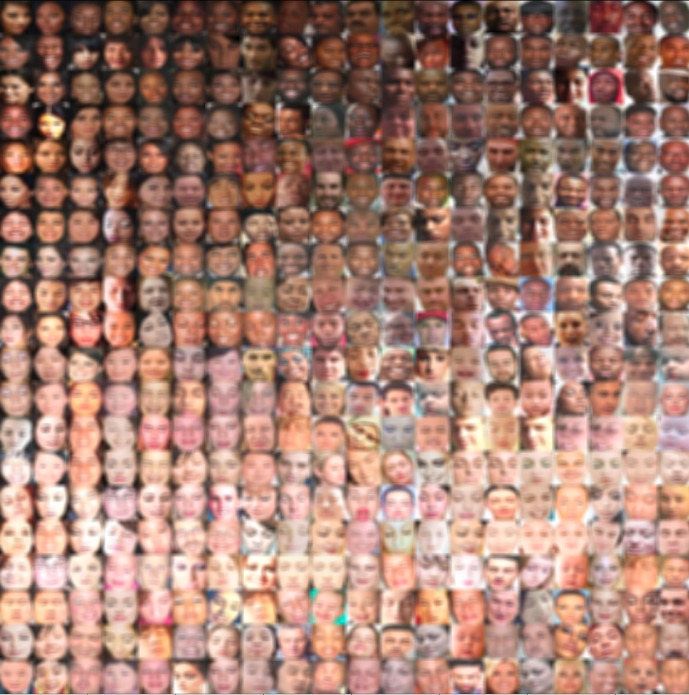}
\caption{The proposed visualisation strategy uses PCA to sort faces by similarity without requiring meta-data. \label{PCA_results}}
\end{figure}

In this work, we retain only two columns ($j=2$), and project images into a two dimensional coordinate system. Figure \ref{grid} shows the 2D projection (coordinates obtained from $\mathbf{V}_{0:2}$) of facial images in our test dataset. We use this projection to construct a grid of images, grouped by similarity. Algorithm \ref{PCA} describes this process. We create a uniform coordinate grid and search for the closest image (in the reduced dimensional coordinate system) to each point in the grid. We assign each image a corresponding point, and ensure that no image is duplicated, by removing it from the list of available images once allocated a grid coordinate in order to produce a grid of images that groups individuals by facial similarity, as shown in Figure \ref{PCA_results}. It is clear that this process successfully groups faces of similar skin tone together, with darker skinned individuals located towards the top of the image, and lighter skinned individuals towards the bottom. For each image selected, we calculate the error in prediction, to produce a saliency map indicating model quality for the constructed grid of images.

\begin{algorithm}
\caption{\textbf{Image overlay generation}}
\label{PCA}
Let $\mathbf{p}$ be the $N\times 2$ matrix $\mathbf{V}_{0:2}$ \\
 \textbf{Input:} $\mathbf{p}$, list of images  $\mathbf{x}_i$, where $i=1\dots N$, $\text{labels}_{i}$
\begin{algorithmic}[1]
\State x-min = min($\mathbf{p_0}$)
\State x-max = max($\mathbf{p_0}$)
\State y-min = min($\mathbf{p_1}$)
\State y-max = max($\mathbf{p_1}$)
\State image-grid = [ ][ ]
\State overlay-grid = [ ][ ]
\State j = 0
\For{pos-x in x-min : d1 : x-max }
\State k = 0
\For{pos-y in y-min : d2 : y-max}
\State min-dist = 10000
\State best-idx = 0
\For{i in range(0,N)}
\State dist = $\sqrt{ (p[i,0]-j)^2 + (p[i,1]-k)^2}$
\If {dist $<$ min-dist}
\State min-dist = dist
\State best-idx = i
\EndIf
\State k = k +1
\State image-grid[j,k] = $\mathbf{x}_i$
\State $\text{overlay-grid}[i,k]=|\text{labels}_{i}-\text{cnn-model}(\mathbf{x}_i)|$
\State remove image $\mathbf{x}_i$ from image-list
\EndFor
\State  j = j + 1
\EndFor
\State \textbf{{Output:}} Returns an overlayed saliency image
\EndFor
\end{algorithmic}
\end{algorithm}

\section{EXPERIMENTAL METHOD}

We examine the potential of CNN-based drowsiness detection algorithms to exhibit algorithmic bias by training a variety of popular classification models on a number of publically available drowsiness detection datasets. A number of strategies were applied to prevent overfitting, so as to ensure a fair analysis.

\subsection{MODEL ARCHITECTURE}

\begin{figure*}[h]
\centering
\includegraphics [width=0.85\textwidth]{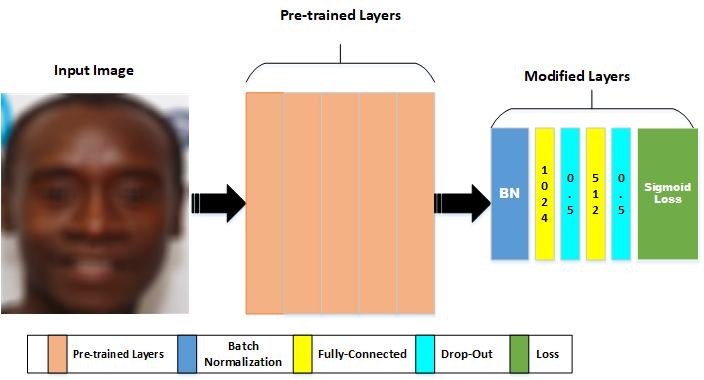}
\caption{The figure shows the neural network architecture used for detection of drowsiness. Pre-trained layers follow the VGG, VGG-face and RestNet architectures, which are used as feature extraction layers. The modified layers are trained to perform drowsiness detection. Dropout layers are used to prevent over-fitting.\label{fig5}}
\end{figure*}

The architectures of the networks used for testing were based on a variety of pre-trained network models (VGG-Face, VGG, and ResNet), with the final layers modified as shown in Figure \ref{fig5}, with Batch Normalisation (BN) applied and three fully connected layers added to the network. 

Pre-trained models (trained for general image classification) were used as feature extractors to lower the number of parameters to be trained and reduce training time. In addition, lower level features are already learned for the pre-trained models, which can prevent overfitting to smaller datasets.  We made use of the Adam optimizer and a binary cross entropy loss in the training process. Data augmentation was also used in an attempt to prevent over-fitting. Here, re-scaling, shearing, zooming, and horizontal flipping was applied to extend the size of datasets used for training. 

Furthermore, zooming was applied to images because the face is of greater interest in drowsiness detection. Horizontal flipping was also applied to generate different angles of the drivers' faces. Dropout ($\alpha=0.5$) was applied between fully connected layers to reduce the chances of over-fitting even further.  

\subsection{DATASETS}

This section describes the datasets that were used for training and testing. For this work, the NTHU-drowsy, DROZY, and CEW datasets are used.

\begin{LaTeXdescription}
\item[NTHU-drowsy] was introduced at the 13th Asian Conference on Computer Vision (ACCV2016) \cite{NTHU}. The dataset is split into test and training sets. For training, there are 18 participants (10 men and 8 women) pretending to drive, with 5 scene scenarios for each participant including no-glasses, glasses, glasses at night, no glasses at night, and sunglasses. For evaluation, there are images of 2 men and 2 women. Videos combining drowsy, normal and sleepy states are provided.

\item[DROZY] consists of 14 participants (3 males and 11 females) \cite{massoz2016ulg}. Each video is approximately 10 minutes long and is accompanied by the results of psychomotor vigilance tests (PVTs) regarding the drowsiness state.  For each participant, the dataset contains a time-synchronized Karolinska Sleepiness Scale (KSS) score \cite{massoz2016ulg}.

\item[CEW] is a collection of online images of different races (for example Asians and non-Asians with light-skinned faces) and contains about 2423 participants \cite{ClosedEyes}. Among the participants, 1192 have both eyes closed and 1231 have their eyes open. These images were selected from the labeled faces in the wild database.

\item[Our Test Dataset] was prepared from a collection of online videos of South African faces. There are 30348 images comprising a variety of different races and ethnicities represented in South Africa. Images range from dark to light-skinned faces of multiple genders to provide a diverse testing dataset.
\end{LaTeXdescription}

The drowsiness detection models were trained and evaluated on these three datasets, which all consist of two classes (awake and drowsy). Three models were trained on each dataset (on over 300k images) and evaluated on 50k of images that were held out from each of the training datasets used. Finally, the South African test dataset was used to test the three trained models.

We prepared all the images from the three datasets in the same manner for training. All images were resized to 150x150 pixels, before applying augmentation and feeding the data in batches into the model. The Adam optimizer was used to train the model and training was performed for 30 epochs. The batch size was kept constant at 32 as it was observed that using a larger batch size degraded the model's quality. 

\section{RESULTS}

\begin{figure}
\centering
\includegraphics [width=0.45\textwidth]{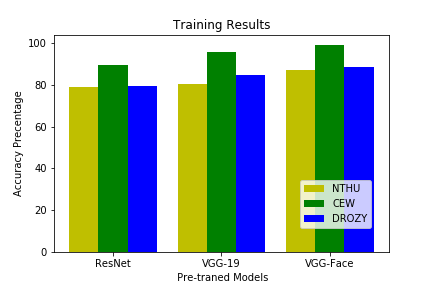}
\caption{The figure shows validation accuracies obtained for each model when testing on data held out from the training datasets. These accuracies seem to indicate that all the models will generalize well.\label{fig1}}
\end{figure}

The accuracy obtained when testing on data held out during training for each dataset is shown in Figure \ref{fig1}. All the training datasets include images of light-skinned individuals, but only the CEW contains images of a small number of dark-skinned individuals. All facial images are blurred for confidentiality.

\begin{figure}
\centering
\includegraphics [width=0.45\textwidth]{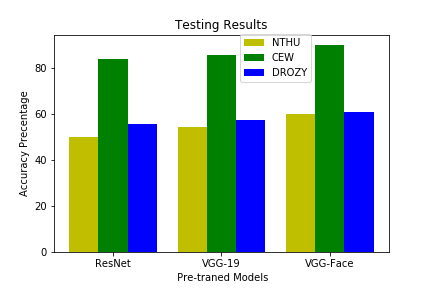}
\caption{When testing on the representative South African dataset, both the NTHU-drowsy and DROZY models failed to generalise, but the CEW model seems to perform well.\label{fig2}}
\end{figure}

The testing dataset was prepared in the same way as the training dataset and using the same parameters for data augmentation.  The loss and accuracy were recorded for both the training and testing phases. Pre-trained models performed well when tested on data held out from training sets. However, all models showed decreased performance when tested on our representative dataset, as shown in Figure \ref{fig2}, although the decrease in performance was marginal for the CEW dataset. It is clear that the models trained using NHTU-drowsy and DROZY completely overfit to these datasets, and failed entirely when tested on our dataset. As a result, the NHTU-drowsy and DROZY are excluded from further analysis.

\begin{figure}[!h]
\centering
\subfloat[]{\includegraphics[width=0.12\textwidth,height=2cm]{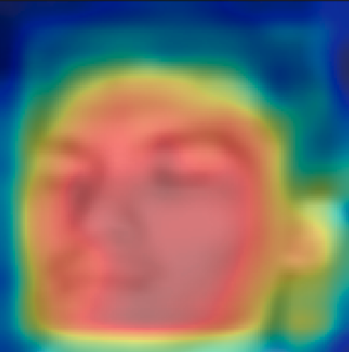}}
\subfloat[]{\includegraphics[width=0.12\textwidth,height=2cm]{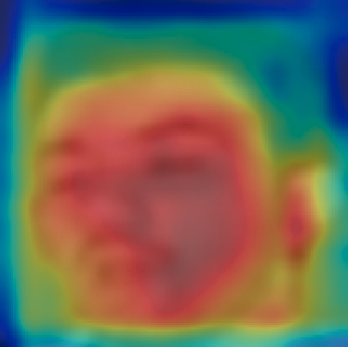}}
\subfloat[]{\includegraphics[width=0.12\textwidth,height=2cm]{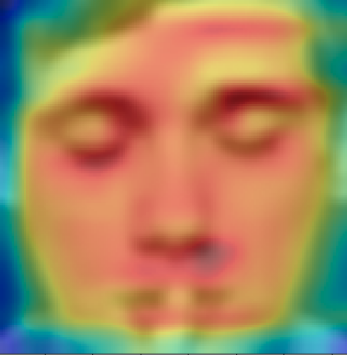}}
\subfloat[]{\includegraphics[width=0.12\textwidth,height=2cm]{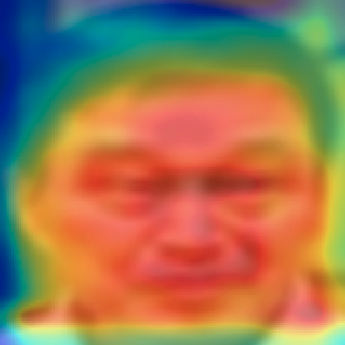}}\\
\subfloat[]{\includegraphics[width=0.12\textwidth,height=2cm]{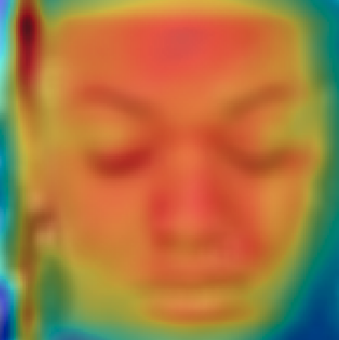}}
\subfloat[]{\includegraphics[width=0.12\textwidth,height=2cm]{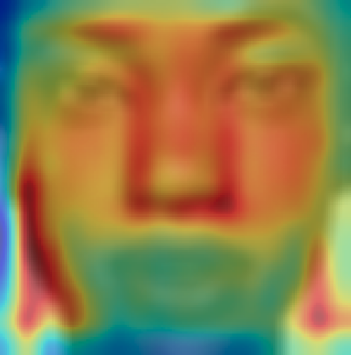}}
\subfloat[]{\includegraphics[width=0.12\textwidth,height=2cm]{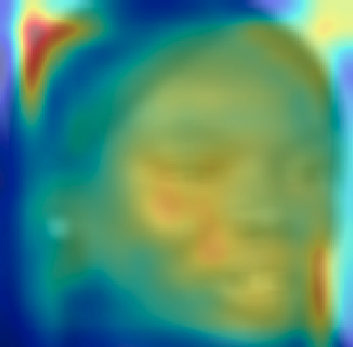}}
\subfloat[]{\includegraphics[width=0.12\textwidth,height=2cm]{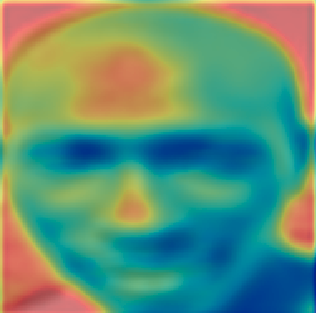}}
\caption{The saliency map overlays highlight pixels in the input image that contribute to the network's final output. Areas marked in red contribute significantly, while blue regions contribute little to the final classification decision. Images (a) to (d) are from the validation set, and the saliency map highlight facial features, as would be expected for a drowsiness detector. Images (e) to (h) were sampled from our test dataset. The saliency visualisation failed to highlight facial features in the images (g) and (h). \label{fig3}}
\end{figure}

Figure \ref{fig3} shows a selection of saliency maps obtained when the VGG-Face models trained using the CEW dataset were tested using both light-skinned individuals (dominant in the training sets) and dark-skinned individuals (dominant in our test set). Here, red areas denote image pixels that contributed significantly to the algorithm output. Interestingly, the model seems to focus on facial regions for the lighter skinned individuals shown here, but fails to do so for darker skinned individuals, indicating a potential failure case.

\begin{figure*}
\centering
\subfloat[ResNet-CEW]{\includegraphics[width=0.34\textwidth,height=6cm]{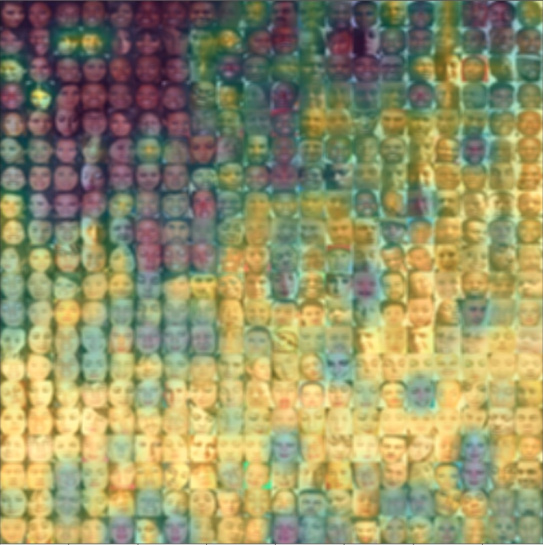}}
\subfloat[VGG-CEW]{\includegraphics[width=0.34\textwidth,height=6cm]{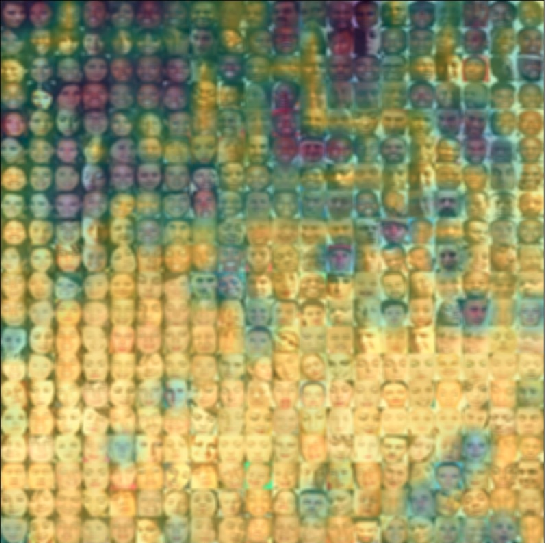}}
\subfloat[VGGFace-CEW]{\includegraphics[width=0.34\textwidth,height=6cm]{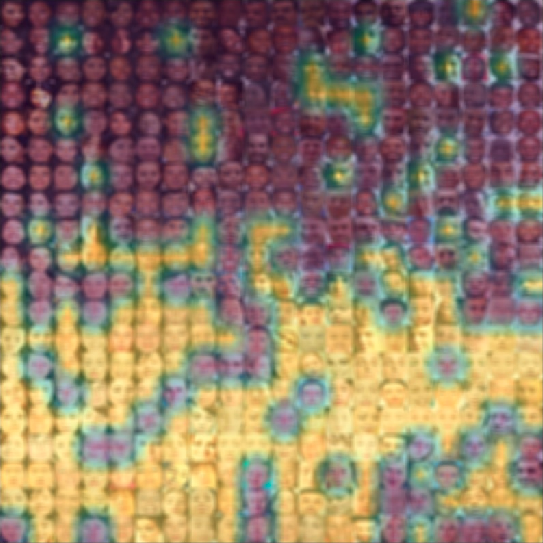}}
\caption{The figure shows images produced using the proposed visualisation strategy. All the trained models appear to be failing on the population groups on the upper part of the image. The yellow shaded parts indicate where the model performs well, while failures are indicated by the purple shaded parts, which appear mostly on the upper part of the images. The green shaded parts show that the model is also performing well, but with lower probability (0.65 to 0.80). \label{fig4}}
\end{figure*}

We also applied the proposed PCA-saliency visualisation strategy (Figure. \ref{fig4}). Although the accuracy measures highlighted previously showed that the CEW models performed well, the proposed visualisation shows that the CEW models seem to struggle to predict drowsiness for darker-skinned individuals at the top of the image, potentially indicating a population group for which additional data is required to train a better model.

The key findings of these experiments are as follows:
\begin{itemize}
\item All training experiments performed well when tested on data held out from training datasets (85.3\% - 98.7\%).
\item All three models showed a decrease in performance when tested on our more representative test dataset, indicating some overfitting. 
\item Models trained using NTHU and DROZY datasets completely failed to generalise.
\item Models trained using the CEW dataset fail to perform well for certain dark-skinned individuals indicating a need for additional training data covering these population groups.
\end{itemize}

\section{CONCLUSION}

Results presented in this paper have showed that models trained using publicly available datasets for drowsiness detection do not generalise well when tested on dark-skinned races. The 50\% accuracy obtained when testing the NTHU and DROZY models on a more representative dataset shows that the network is simply guessing the drowsiness state of the driver, which could lead to system failure and endanger drivers if these models were deployed. In contrast, the CEW models appear to perform well, but further examination shows that they are failing systematically on certain population groups. 
 
This paper has highlighted the potential for racial discrimination by machine learning models when the datasets used for training do not cover the demographics present where the system might be deployed. Going forward, it is crucial that balanced training datasets, covering all races and ethnicities, are used to train systems for driver aid. This work has introduced a visualisation strategy that can be used to identify population groups on which an algorithm is failing, without the need for meta-data regarding race or ethnicity. Furthermore, this work has shown that there is a strong need to evaluate vision-based driver drowsiness detection systems in the countries where they will be deployed, in order to prevent unintentional discrimination. 

\bibliographystyle{IEEEtran}  
\bibliography{refs}
\end{document}